\documentclass[letterpaper, 10 pt, conference]{ieeeconf}  

\IEEEoverridecommandlockouts                              

\usepackage{setspace}
\usepackage{cite}
\usepackage{multicol}
\usepackage[hidelinks]{hyperref}
\usepackage{comment}
\usepackage{amrl}
\usepackage{booktabs}
\usepackage{multirow}
\usepackage{graphicx}
\usepackage{subcaption}
\usepackage{caption}
\usepackage{xcolor}
\usepackage{booktabs}
\usepackage{multirow}
\usepackage{siunitx}
\usepackage{float}

\usepackage[table]{xcolor}
\definecolor{myorange}{HTML}{E97132}

\usepackage{amsmath, amssymb}
\usepackage{bbm}
\newcommand{\policyname}{\texttt{VOFA}}

\begin{document}


\title{\huge \bf \policyname{}: Visual Object Goal Pushing with Force-Adaptive Control for Humanoids}

\author{ 
\textbf{Zichao Hu\textsuperscript{1}\thanks{Correspondence to: \texttt{zichao@utexas.edu}}, Zifan Xu\textsuperscript{1}, Dongsik Chang\textsuperscript{2}, He Yin\textsuperscript{2}, Linh Tran\textsuperscript{1}, Roberto Mart\'{\i}n-Mart\'{\i}n\textsuperscript{1,2}}\\
\textbf{Peter Stone\textsuperscript{1,3}, Jingyu Qiao\textsuperscript{2}, Joydeep Biswas\textsuperscript{1}}\\[3pt]
\textsuperscript{1}Department of Computer Science, The University of Texas at Austin\\
\textsuperscript{2} Amazon Inc.
\textsuperscript{3}Sony AI 
}

\IEEEaftertitletext{{
            \vspace{-4mm}
            \centering
        \includegraphics[width=0.99\linewidth]{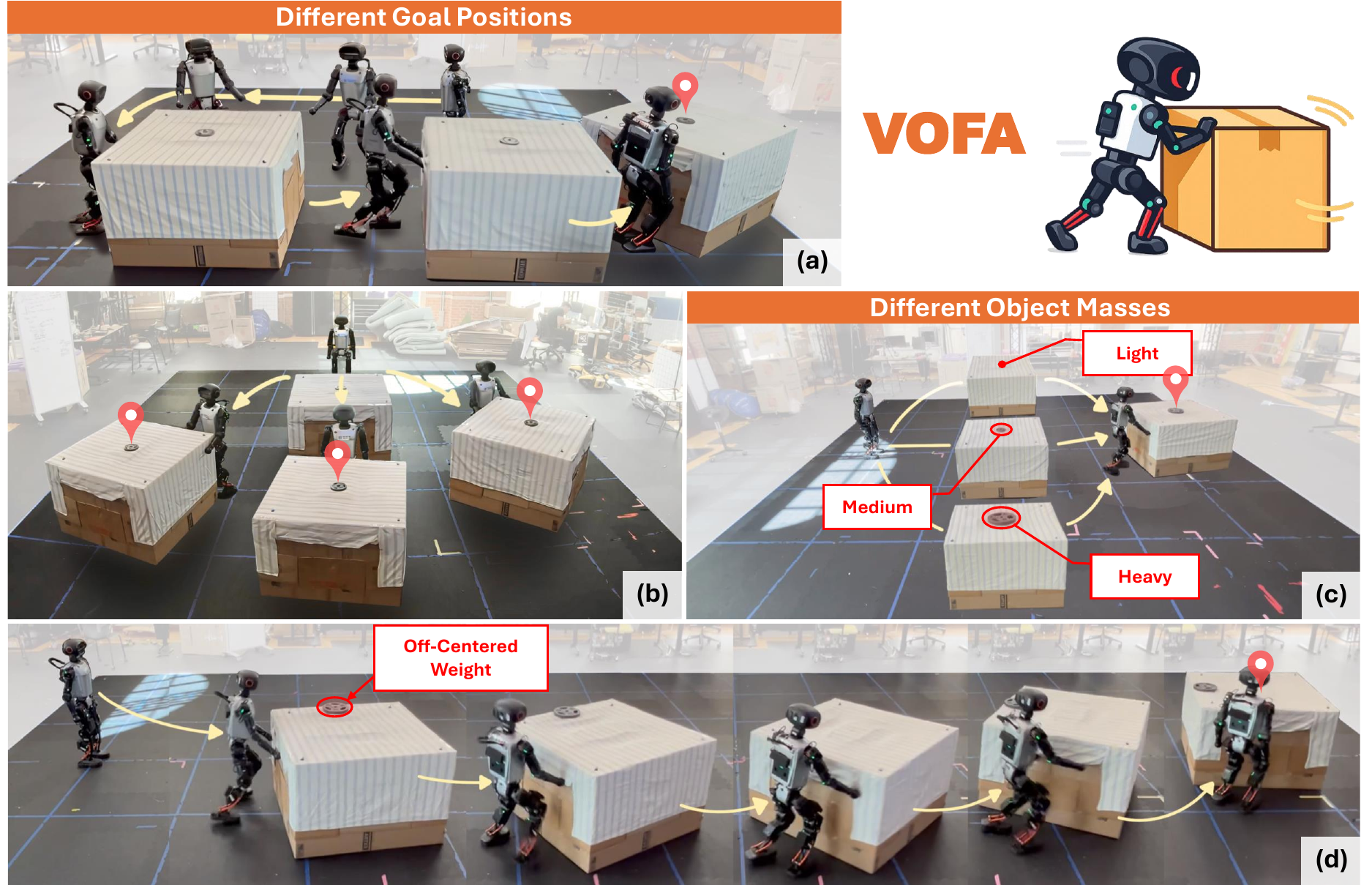}
            \captionof{figure}{We present \policyname{}, a visual goal-conditioned humanoid loco-manipulation system capable of pushing objects with unknown physical properties. The system adapts to different goal positions (a,b), object masses (c), and center-of-mass configurations (d) while maintaining stable, closed-loop control.}
            \label{fig:first_figure}
        }}
\maketitle  

\begin{abstract}
The ability to push large objects in a goal-directed manner using onboard egocentric perception is an essential skill
for humanoid robots to perform complex tasks such as material handling in warehouses.
To robustly manipulate heavy objects to arbitrary goal configurations, the robot must 
cope with unknown object mass and ground friction, noisy onboard perception, and actuation errors; all in a real-time feedback loop.
Existing solutions either rely on privileged object-state information without onboard perception or lack robustness to variations in goal configurations and object physical properties.
In this work, we present \policyname{}, a visual goal-conditioned humanoid loco-manipulation system capable of pushing objects with unknown physical properties to arbitrary goal positions. 
\policyname{} consists of a two-level hierarchical architecture with a high-level visuomotor policy and a low-level force-adaptive whole-body controller. The high-level policy processes noisy onboard observations and generates goal-conditioned commands to operate in closed loop across diverse object–goal configurations, while the low-level whole-body controller provides robustness to variations in object physical properties.
\policyname{} is extensively evaluated in both simulation and real-world experiments on the Booster T1 humanoid robot. Our results demonstrate strong performance, achieving over 90\% success in simulation and over 80\% success in real-world trials. Moreover, \policyname{} successfully pushes objects weighing up to 17kg, exceeding half of the Booster T1’s body weight.
\footnote{Code is available at \href{https://github.com/ut-amrl/VOFA} {https://github.com/ut-amrl/VOFA}.}

\end{abstract}

\IEEEpeerreviewmaketitle


\section{Introduction}

Developing loco-manipulation capabilities for humanoid robots to perform useful tasks has gained increasing attention in robotics research~\cite{humanoid_current_progress}. One such capability is pushing large objects to designated locations to support real-world applications such as warehouse inventory reorganization.
Successfully achieving this capability requires addressing several key challenges: (1) handling objects with unknown and diverse physical properties, such as mass and friction; (2) operating with noisy onboard visual observations without access to privileged object-state information; and (3) adapting in closed loop to different object–goal configurations. 
Recent progress in humanoid loco-manipulation driven by reinforcement learning has demonstrated promising capabilities; however, existing humanoid systems do not fully address the challenges presented by goal-directed object pushing. In particular, existing systems rely on privileged object-state sensing~\cite{quadruped_pushing1, quadruped_pushing2, r2s2, hdmi, hitter}, lack adaptability to different object and goal configurations~\cite{visualmimic, humanoid_open_door, zhi2025learningunifiedforceposition}, or do not explicitly account for variations in object mass~\cite{viral, slac}.

\begin{figure*}[t]
    \centering
    \includegraphics[width=\linewidth]{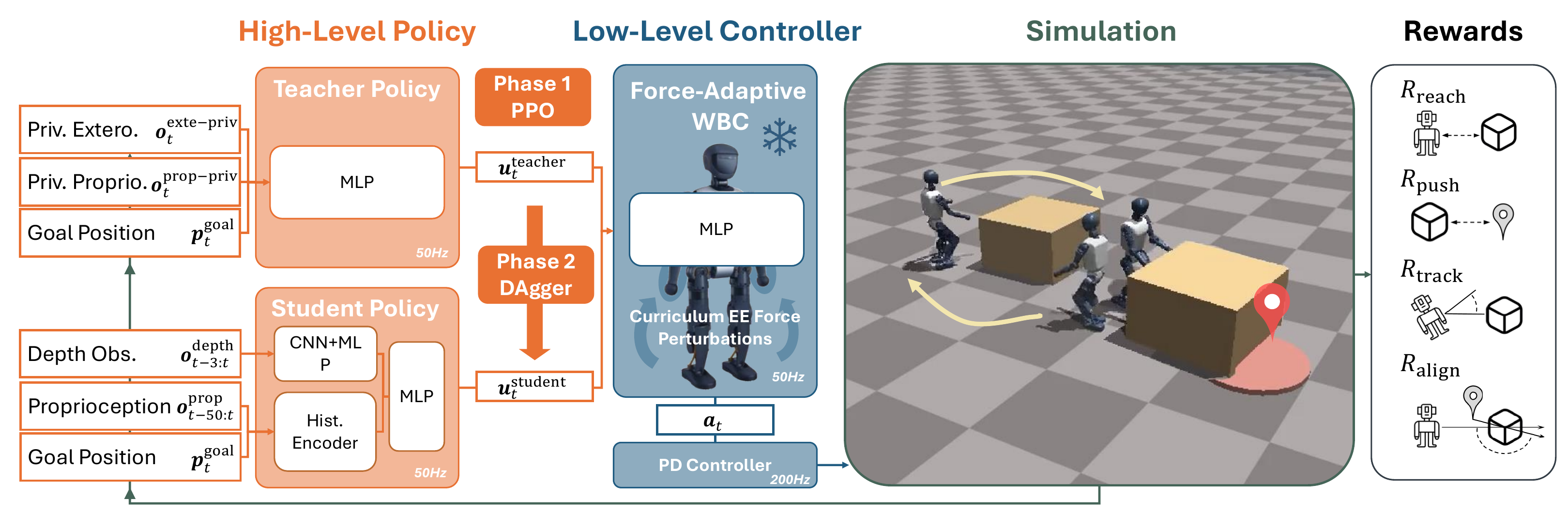}
\caption{\textbf{\policyname{} Design.}
\policyname{} adopts a hierarchical architecture that combines a high-level visuomotor policy with a force-adaptive whole-body controller~\cite{falcon} for humanoid visual object–goal pushing. The high-level policy is trained in a teacher–student framework: a teacher policy is trained with privileged observations and goal positions using PPO~\cite{ppo}, and a vision-based student policy is distilled via DAgger~\cite{dagger} using onboard sensory inputs and goal information. Training uses four reward terms—reach, push, track, and alignment—to promote robust closed-loop object–goal interaction.}
    \label{fig:main-method}
\end{figure*}

In this work, we present \policyname{} (\texttt{V}isual \texttt{O}bject–Goal Pushing with \texttt{F}orce-\texttt{A}daptive Control), a visual goal-conditioned humanoid loco-manipulation system capable of pushing objects with unknown physical properties (e.g., mass and friction) to arbitrary goal positions. \policyname{} employs a two-level hierarchical architecture (\figref{main-method}) composed of a high-level depth-image-based visuomotor policy and a low-level force-adaptive whole-body controller (WBC). The high-level policy processes onboard observations and outputs goal-conditioned commands, which are translated by the low-level controller into joint-level control actions. 
To address unknown object physical properties, the low-level controller is explicitly designed as a force-adaptive controller following the FALCON~\cite{falcon} framework. This controller enables stable object interaction across diverse object masses and ground frictions. Together, the hierarchical design decouples low-level force-adaptive control from high-level visuomotor planning, simplifying policy training~\cite{visualmimic} while preserving robustness to variations in object physical properties.

To efficiently learn a high-level policy driven by noisy onboard visual observations, \policyname{} follows a teacher–student design. The teacher policy is first trained as a goal-conditioned policy using privileged proprioceptive and exteroceptive information (\eg{} the robot’s base linear velocity and the object’s orientation). DAgger~\cite{dagger} is then employed to distill a vision-based student policy that replaces privileged inputs with proprioceptive signals and onboard depth observations. During distillation, visual augmentations including camera extrinsic randomization and depth-noise injection (\figref{depth_side_by_side}) are incorporated to improve robustness to perception noise and background distractions at deployment.

To enable \policyname{} to adapt in closed loop to arbitrary goal positions, training is conducted in large-scale Isaac Gym simulation with extensive randomization over initial object and goal placements. An object–goal alignment reward is further introduced to encourage the robot to first reposition itself to the opposite side of the object relative to the goal before initiating contact (\figref{align_reward_ablation}). This simple yet effective reward design removes the need to manually define separate training stages~\cite{kick_ball, humanoid_open_door} to complete the full task. It promotes consistent object–robot alignment and enables corrective behavior when deviations from the desired object trajectory occur.

Our system, \policyname{}, achieves over 90\% success rates in simulation and over 80\% success rates in real-world experiments across a range of initial object and goal positions and diverse object physical properties. In real-world deployment (\figref{first_figure}), \policyname{} successfully pushes objects weighing 17 kg (more than half of the Booster T1 robot’s body weight) despite not being trained on such object masses in simulation. Leveraging closed-loop visual feedback, \policyname{} remains robust even when the object’s center of mass is intentionally shifted to previously unseen locations, and is able to detect and recover from deviations in the object’s motion online. In addition, \policyname{} supports long-horizon box-pushing tasks, sequentially pushing the object to multiple goal locations. Together, these results demonstrate robust closed-loop performance and highlight the potential of \policyname{} to enable more complex downstream applications, such as object rearrangement. 
Overall, our key contributions are:
\begin{enumerate}
    \item We present a visual humanoid object–goal pushing system that enables closed-loop, goal-directed pushing across diverse object–goal configurations.
    \item Our system handles objects with diverse and even previously unseen masses, successfully pushing objects up to 17kg—more than half of the robot’s body weight.
    \item We validate our approach in both large-scale simulation and real-world experiments on the Booster T1 humanoid robot, achieving high success rates across diverse scene configurations and object physical properties.
\end{enumerate}

\section{Related Work}
In this section, we review related work on humanoid whole-body control and autonomous legged loco-manipulation.

\subsection{Humanoid Whole Body Control}

A substantial body of recent work has explored learning-based whole-body control (WBC) for humanoid robots, demonstrating stable and expressive full-body behaviors such as locomotion, balance, and dynamic motion generation~\cite{asap, exbody2, beyondmimic, hub, sonic}. Many of these approaches rely on large-scale motion imitation and primarily focus on reproducing coordinated human motions. Beyond expressive whole-body motions, other work investigates WBC for teleoperated loco-manipulation, enabling humanoids to interact with objects while maintaining whole-body stability~\cite{amo, clone, falcon, homie, omnih2o, twist}. Among these works, FALCON~\cite{falcon} introduces a force-adaptive whole-body controller for humanoid loco-manipulation, providing robustness to substantial and unknown external forces at the end-effector during object interaction. 
In this work, we build on FALCON and extend it with a hierarchical, vision-based framework to enable goal-conditioned visual object pushing for humanoid robots.



\subsection{Sim-to-Real Reinforcement Learning for Autonomous Humanoid Loco-Manipulation}
Enabling legged robots to autonomously perform loco-manipulation is important for deploying robots to perform useful tasks in human-inhabited environments. 
Prior work on autonomous loco-manipulation has studied both quadruped~\cite{quadruped_pushing1, quadruped_pushing2} and humanoid~\cite{hdmi, r2s2, hitter} robots; however, these approaches typically rely on privileged object state information rather than onboard visual perception.
More recently, vision-based humanoid loco-manipulation has been explored by incorporating onboard visual perception into whole-body policies \cite{visualmimic, viral, humanoid_open_door}; however, these approaches do not explicitly study goal-conditioned object interaction or robustness to variations in object physical properties. 
Visual, goal-directed humanoid loco-manipulation under variations in object mass and physical properties is still challenging. In this work, we aim to address this gap through sim-to-real reinforcement learning.



\section{Method}
\textbf{Problem Setup:} We consider the problem of visual goal-directed object pushing with a humanoid robot. At each timestep $t$, the robot receives observations consisting of proprioceptive measurements $\boldsymbol{o}_t^{\text{prop}}$, depth observations $\boldsymbol{o}_t^{\text{depth}}$, and a desired object goal position $\boldsymbol{p}_t^{\text{goal}}$. The objective is to learn a control policy that maps a history of observations to humanoid joint position targets $\boldsymbol{a}_t$ such that it controls the robot to push the object to the specified goal location.

\textbf{System Overview:} We present \policyname{}, a visual goal-conditioned loco-manipulation system with a two-level hierarchical architecture. The high-level policy
$\pi_H(\boldsymbol{u}_t \mid \boldsymbol{o}_{t-h_1:t}^{\text{prop}}, \boldsymbol{o}_{t-h_2:t}^{\text{depth}}, \boldsymbol{p}_t^{\text{goal}})$
first maps a history of proprioceptive and depth observations and the goal position to a high-level command $\boldsymbol{u}_t$ (\secref{method_high_level_policy_training}). 
Then, the high-level command is provided as input to a low-level force-adaptive controller
$\pi_L(\boldsymbol{a}_t \mid \boldsymbol{u}_t)$,
which outputs joint position targets $\boldsymbol{a}_t$ (\secref{method_low_level_policy_training}). 
Both policies are trained in IsaacGym simulation and operate at 50Hz. 
Beyond the architectural design, we detail the reward formulation (\secref{method_reward_design}) and the domain randomization strategy enabling zero-shot transfer to the real robot (\secref{method_domain_rand}) in the following subsections as well.

\subsection{High-Level Policy Design and Training}
\label{sec:method_high_level_policy_training}
To efficiently train the high-level policy, we adopt a teacher--student paradigm.

\paragraph{Teacher Policy Design}
We first train a goal-conditioned teacher policy to output actions within the low-level WBC's valid command range~\cite{visualmimic}. The teacher policy is trained using privileged observations with PPO~\cite{ppo} under an asymmetric actor–critic framework~\cite{asymmetric_ac}, which has been shown to be effective in prior work~\cite{quadruped_pushing2}.

The actor receives privileged observations $\boldsymbol{o}_t^{\text{priv}} = (\boldsymbol{o}_t^{\text{prop-priv}}, \boldsymbol{o}_t^{\text{exte-priv}}, \boldsymbol{p}_t^{\text{goal}})$, consisting of privileged proprioceptive information, privileged exteroceptive information, and the goal position. Proprioceptive inputs are given by $\boldsymbol{o}_t^{\text{prop-priv}} = (\boldsymbol{v}_t, \boldsymbol{\omega}_t, \boldsymbol{g}_t, \boldsymbol{u}^\text{teacher}_{t-1}, \boldsymbol{q}_t^{\text{upper}}, \dot{\boldsymbol{q}}_t^{\text{upper}})$, including linear and angular velocities, projected gravity, the previous high-level action, upper-body joint positions and velocities. Exteroceptive inputs are given by $\boldsymbol{o}_t^{\text{exte-priv}} = (\boldsymbol{p}_t^{\text{ee,r-obj}}, \boldsymbol{p}_t^{\text{ee,l-obj}}, \boldsymbol{R}_t)$, comprising the relative positions of the right and left end effectors with respect to the object and the object orientation. All quantities are expressed in the robot base frame.

The critic observes all actor inputs and additionally receives privileged object state information
\(
\boldsymbol{o}_t^{\text{priv-critic}} = \big( c_t,\; \boldsymbol{p}_t^{\text{obj}},\; m_t^{\text{obj}},\; \boldsymbol{d}_t^{\text{obj}},\; \boldsymbol{v}_t^{\text{obj}},\; \boldsymbol{\omega}_t^{\text{obj}} \big),
\)
where $c_t$ denotes the object contact flag, $\boldsymbol{p}_t^{\text{obj}}$ is the object center-of-mass position, $m_t^{\text{obj}}$ and $\boldsymbol{d}_t^{\text{obj}}$ represent the object mass and dimensions, and $\boldsymbol{v}_t^{\text{obj}}$ and $\boldsymbol{\omega}_t^{\text{obj}}$ are the object linear and angular velocities. 

\paragraph{Student Policy Design}

We employ DAgger~\cite{dagger} to distill a vision-based student policy from the teacher by replacing privileged proprioceptive and exteroceptive inputs $(\boldsymbol{o}_t^{\text{prop-priv}}, \boldsymbol{o}_t^{\text{exte-priv}})$ with a history of onboard proprioceptive observations and depth images $(\boldsymbol{o}_{t-50:t}^{\text{prop}}, \boldsymbol{o}_{t-3:t}^{\text{depth}})$. The proprioceptive observation $\boldsymbol{o}_t^{\text{prop}} = (\boldsymbol{\omega}_t, \boldsymbol{g}_t, \boldsymbol{u}_{t-1}^{\text{student}}, \boldsymbol{q}_t, \dot{\boldsymbol{q}}_t)$ consists of the base angular velocity, projected gravity, the previous high-level action, and the joint positions and velocities.
The student policy encodes the proprioceptive history using temporal 1D convolutions~\cite{kick_ball}, while each depth observation is processed by a convolutional neural network to extract visual features. The proprioceptive and visual embeddings are then fused and passed through an MLP projection head that aggregates the combined representation for action prediction.
To improve training efficiency, the policy maintains a rolling buffer of the three most recent depth frames, processes depth observations at 5Hz, and downsamples the resolution of images (from 640$\times$360 to 32$\times$32).

\subsection{Low-Level Policy Design and Training}
\label{sec:method_low_level_policy_training}
We adopt a force-adaptive whole-body controller (WBC) as the low-level policy to handle substantial and unknown end-effector interaction forces, which commonly arise in object-pushing tasks due to variations in object physical properties (e.g., mass). 
In this work, we choose FALCON~\cite{falcon} as it is explicitly trained for force-adaptive control using a curriculum that progressively increases external forces at the end-effectors. 
In addition, FALCON is trained on the AMASS~\cite{AMASS} motion dataset, which spans a broad range of natural human motions, including behaviors relevant to box pushing.
FALCON receives a high-level command $\boldsymbol{u}_t = (\boldsymbol{u}_t^{\text{lower}}, \boldsymbol{q}_t^{\text{upper}})$, where the lower-body locomotion command 
$\boldsymbol{u}_t^{\text{lower}} = (\boldsymbol{v}_t^{\mathrm{lin,ang}}, \phi_t^{\mathrm{stance}}, h_t^{\mathrm{root}}, \omega_t^{\mathrm{yaw}})$ 
specifies the desired root linear and angular velocities, walking-mode indicator, root height, and waist yaw angle, and $\boldsymbol{q}_t^{\text{upper}}$ denotes upper-body target joint positions. In this work, the walking-mode indicator, root height, and waist yaw angle are set to their default values ($1$, $0.62$, and $0$, respectively), while learning a residual upper-body target joint position $\Delta \boldsymbol{q}_{t}^{\text{upper}}$. FALCON is trained for 10k epochs using PPO and is kept frozen during subsequent high-level policy training.


\subsection{Reward Design}
\label{sec:method_reward_design}

To enable the teacher policy to learn effective object goal pushing behavior, we design four main rewards:

\begin{enumerate}
    \item \textbf{Reaching the object with both end-effectors.}
    To encourage both end-effectors to approach the object, we define the per–end-effector reaching terms as
    \begin{equation}
    r^{\text{ee}_i}_t
    = \exp\!\left(\frac{-\lVert \boldsymbol{p}_t^{\text{ee,i}-\text{obj}} \rVert^2}{\sigma^2_{\text{reach}}}\right), \quad i\in\{l,r\}.
    \end{equation}

    The overall reaching reward is computed as the harmonic mean of the two end-effector terms:
    \begin{equation}
    R_{\text{reach}} = \mathrm{HarmonicMean}\!\left(r^{\text{ee}_l}_t,\, r^{\text{ee}_r}_t\right).
    \end{equation}

    \item \textbf{Pushing the object toward the goal.}
    The robot is rewarded for reducing the object–goal distance:
    \begin{equation}
        R_{\text{push}} =
        \exp\!\left(\frac{-\|\boldsymbol{p}_t^{\text{goal}} - \boldsymbol{p}_t^{\text{obj}}\|^2}{\sigma_\text{push}^2}\right).
    \end{equation}

 \item \textbf{Head tracking reward.}
    The robot is encouraged to visually track the object:
    \begin{equation}
        R_{\text{track}} =
        \begin{cases}
            \exp\!\left(\frac{-\theta_t^2}{\sigma^2_\text{track}}\right), & \text{if } |\theta_t| \le \theta_{\text{fov}}, \\
            0, & \text{otherwise},
        \end{cases}
    \end{equation}
    where $\theta_t$ is the angular deviation between the head (camera optical axis) and the object direction, and $\theta_{\text{fov}}$ denotes half of the camera field of view. 
    \item \textbf{Object–goal alignment reward.}
    To promote goal-directed pushing configurations, we define:
    \begin{equation}
        R_{\text{align}} =
        \exp\!\left(\frac{-(1 - \theta_t^{\text{align}})^2}{\sigma^2_\text{align}}\right),
    \end{equation}
    where $\theta_t^{\text{align}}$ is the normalized angle between the object–robot and object–goal direction vectors.
\end{enumerate}

Notably, the object–goal alignment reward is critical for learning effective goal-conditioned pushing behavior, as demonstrated by the ablation study in~\secref{align_reward}. In addition to these four main task rewards, we incorporate several penalty terms to regularize the robot’s behavior and improve stability. 


\begin{table*}[t]
\centering
\caption{Performance under different initial goal positions.}
\label{tab:object_goal_configuration_exp}

\renewcommand{\arraystretch}{1.15}
\setlength{\tabcolsep}{3.5pt}

\resizebox{\linewidth}{!}{
\begin{tabular}{
l
S
S S S
S S S
S S S
}
\toprule

& {Overall}
& \multicolumn{3}{c}{Front}
& \multicolumn{3}{c}{Lateral}
& \multicolumn{3}{c}{Rear} \\

\cmidrule(lr){2-2}
\cmidrule(lr){3-5}
\cmidrule(lr){6-8}
\cmidrule(lr){9-11}

Method
& {SR(\%) $\uparrow$}
& {SR(\%) $\uparrow$} & {Pre-C.(\%) $\downarrow$} & {Post-C. $\downarrow$}
& {SR (\%) $\uparrow$} & {Pre-C.(\%) $\downarrow$} & {Post-C.(\%) $\downarrow$}
& {SR(\%) $\uparrow$} & {Pre-C.(\%) $\downarrow$} & {Post-C.(\%) $\downarrow$} \\

\midrule
\rowcolor{myorange!25}
Teacher w/ FA (ours)
& \bfseries 97.84
& \bfseries 98.12 & \bfseries 0.72 & \bfseries 0.96
& \bfseries 97.58 & \bfseries 1.00 & \bfseries 1.16
& \bfseries 97.34 & \bfseries 1.12 & \bfseries 1.08 \\

Teacher w/o FA
& 76.94
& 83.78 & 2.92 & 5.52
& 79.00 & 4.88 & 6.58
& 71.86 & 6.46 & 7.38 \\

\midrule
\rowcolor{myorange!25}
Student w/ FA (ours)
& \bfseries 91.38
& \bfseries 91.97 & \bfseries 1.84 & \bfseries 2.42
& \bfseries 91.40 & \bfseries 1.30 & \bfseries 3.22
& \bfseries 90.22 & \bfseries 1.92 & \bfseries 3.98 \\

Student w/o FA 
& 59.32
& 71.16 & 2.68 & 7.00
& 64.90 & 4.52 & 7.78
& 56.06 & 5.64 & 8.84 \\

\bottomrule
\end{tabular}
}
\end{table*}

\begin{table*}[t]
\centering
\caption{Performance under different box mass.}
\label{tab:object_mass_configuration}

\renewcommand{\arraystretch}{1.15}
\setlength{\tabcolsep}{3pt}

\resizebox{\linewidth}{!}{
\begin{tabular}{
l
S S S
S S S
S S S
S S S
}
\toprule
& \multicolumn{3}{c}{Light}
& \multicolumn{3}{c}{Medium}
& \multicolumn{3}{c}{Heavy}
& \multicolumn{3}{c}{Extra Heavy (Unseen)}\\

\cmidrule(lr){2-4} 
\cmidrule(lr){5-7} 
\cmidrule(lr){8-10} 
\cmidrule(lr){11-13}

Method
& {SR(\%) $\uparrow$} & {Pre-C.(\%) $\downarrow$} & {Post-C.(\%) $\downarrow$}
& {SR(\%) $\uparrow$} & {Pre-C.(\%) $\downarrow$} & {Post-C.(\%) $\downarrow$}
& {SR (\%) $\uparrow$} & {Pre-C.(\%) $\downarrow$} & {Post-C.(\%) $\downarrow$}
& {SR(\%) $\uparrow$} & {Pre-C.(\%) $\downarrow$} & {Post-C.(\%) $\downarrow$} \\

\midrule
\rowcolor{myorange!25}
Teacher w/ FA (ours)       
& \bfseries 97.40 & \bfseries 1.12 & \bfseries 1.34 
& \bfseries 98.22 & \bfseries 0.92 & \bfseries 0.74 
& \bfseries 97.46 & \bfseries 1.06 & \bfseries 1.08 
& \bfseries 87.92 & \bfseries 0.88 & \bfseries 4.16 \\

Teacher w/o FA       
& 82.18 & 5.40 & 9.00 
& 80.98 & 5.08 & 6.06 
& 69.76 & 4.64 & 6.06 
& 53.68 & 4.78 & 5.66 \\

\midrule
\rowcolor{myorange!25}
Student w/ FA (ours)    
& \bfseries 93.50 & \bfseries 1.46 & \bfseries 3.06 
& \bfseries 93.30 & \bfseries 1.40 & \bfseries 2.74 
& \bfseries 89.16 & \bfseries 1.22 & \bfseries 3.22 
& \bfseries 69.00 & \bfseries 1.30 & \bfseries 6.30    \\

Student w/o FA
& 70.98 & 6.60 & 9.46
& 62.68 & 5.98 & 7.64 
& 48.66 & 6.56 & 9.58 
& 32.00 & 5.36 & 9.94    \\
\bottomrule
\end{tabular}
}
\end{table*}

\subsection{Environment Setup and Domain Randomization}
\label{sec:method_domain_rand}

\begin{figure}[t]
    \centering
    \begin{subfigure}{0.32\linewidth}
        \centering
        \includegraphics[width=\linewidth]{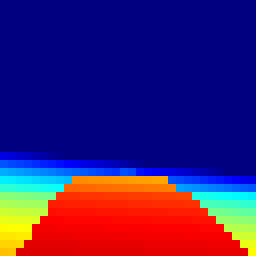}
        \caption{Sim Vanilla}
        \label{fig:depth_raw}
    \end{subfigure}
    \hfill
    \begin{subfigure}{0.32\linewidth}
        \centering
        \includegraphics[width=\linewidth]{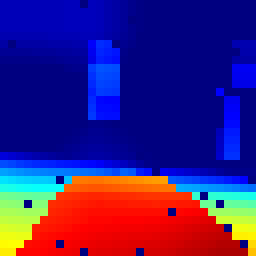}
        \caption{Sim w/ Aug.}
        \label{fig:depth_processed}
    \end{subfigure}
    \hfill
    \begin{subfigure}{0.32\linewidth}
        \centering
        \includegraphics[width=\linewidth]{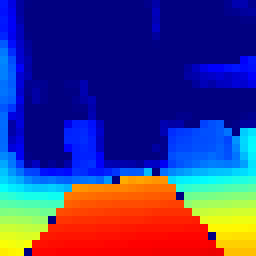}
        \caption{Real}
        \label{fig:real}
    \end{subfigure}
    \caption{
    \textbf{Egocentric vision of the humanoid robot}. 
    During training, we apply three vision augmentations:
    \textit{far-plane depth perturbation}, \emph{correlated depth noise}, and \emph{pixel dropout}.
    }
    \label{fig:depth_side_by_side}
\end{figure}

During training, we randomize object physical properties and initial object–goal configurations. We consider a box-shaped object and randomize its mass (1–8 kg) and friction coefficients (0.1–1.0) to promote robustness across diverse physical conditions. Goals are placed uniformly over a $360^\circ$ arc around the robot at a radius of 2.5–3.5 m. Objects are randomly initialized within the robot’s field of view ($\pm 60^\circ$) and initial distances of 1.8–2.2 m.

To reduce the sim-to-real gap in visual perception, we match the camera intrinsics in simulation to those of the real robot’s cameras (ZED 2i). However, small differences in camera mounting can result in slight viewpoint variations, even across robots of the same model. To account for this variability, we apply random extrinsic perturbations during training, including translations of $\pm3$ cm in x,y,z and rotations of $\pm5$° (yaw), $\pm5$° (pitch), and $\pm2$° (roll).
In addition, real-world depth images exhibit sensor artifacts such as interference from background objects and isolated pixels with spuriously large depth values, as shown in \figref{depth_processed}. To emulate these effects, we apply three visual randomizations as shown in \figref{real}. First, we introduce \textit{far-plane depth perturbation}, where random rectangular patches are placed in the upper image region (above an estimated ground boundary) and overwritten with blocky far-depth noise to simulate transient background clutter.
Second, we apply \textit{correlated depth noise} by adding spatially correlated low-frequency Gaussian noise to the depth map, modeling structured sensor distortions.
Third, we apply \textit{pixel dropout}, randomly zeroing depth pixels to simulate missing or unreliable returns. An ablation study on the effectiveness of these randomizations is presented in \secref{effectiveness_visual_randomization}.

\begin{figure*}[t]
    \centering
    \begin{subfigure}[t]{0.48\linewidth}
        \centering
        \includegraphics[width=\linewidth]{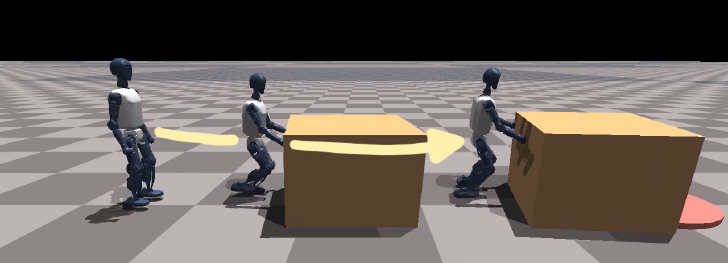}
        \caption{w/ Force-Adaptive Controller (ours)}
        \label{fig:w_align}
    \end{subfigure}
    \begin{subfigure}[t]{0.48\linewidth}
        \centering
        \includegraphics[width=\linewidth]{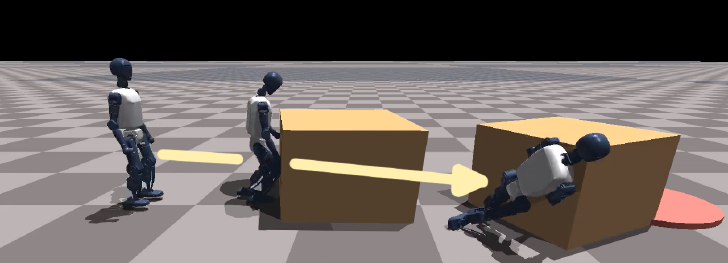}
        \caption{w/o Force-Adaptive Controller}
        \label{fig:wo_align}
    \end{subfigure}
    \caption{\textbf{Force-Adaptive Controller Ablation.}
With the force-adaptive controller, the robot stably pushes the object using its end-effector (a). Without it, the robot struggles to apply consistent forces, often resorting to kicking, which results in unstable behavior and a higher risk of falling.}
    \label{fig:force_adaptive_ablation}
\end{figure*}

\section{Experimental Results}
In this section, we conduct a series of simulation and real-world experiments, with policies trained in IsaacGym and deployed on the Booster T1 humanoid robot, to answer the following research questions:
\begin{enumerate}
    \item How robust is \policyname{} across varying initial goal configurations and object masses (\secref{performance_quantitative_simulation})? 
    \item Is using a force-adaptive whole-body control beneficial for \policyname{} (\secref{effect_of_fa_control})?
    \item Does the object–goal alignment reward design mitigate failure modes in challenging initial configurations (\secref{align_reward})? 
    \item How well does the policy transfer from simulation to the real world (\secref{sim2real_performance})?
    \item How effective is the proposed visual randomization strategy for sim-to-real transfer (\secref{effectiveness_visual_randomization})?
    \item Can the policy perform with closed-loop control when pushing the object (\secref{closed_loop_behavior})?
\end{enumerate}

\begin{figure*}[t]
    \centering
    \begin{subfigure}[t]{0.48\linewidth}
        \centering
        \includegraphics[width=\linewidth]{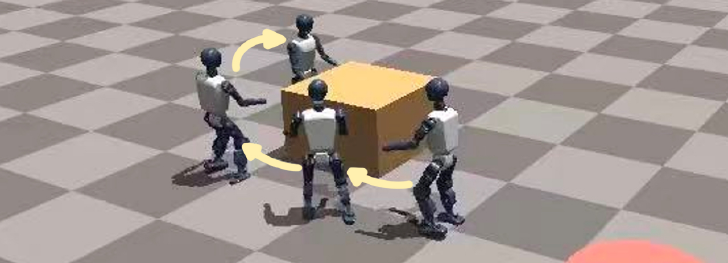}
        \caption{w/ Align Reward (ours)}
        \label{fig:w_align}
    \end{subfigure}
    \begin{subfigure}[t]{0.48\linewidth}
        \centering
        \includegraphics[width=\linewidth]{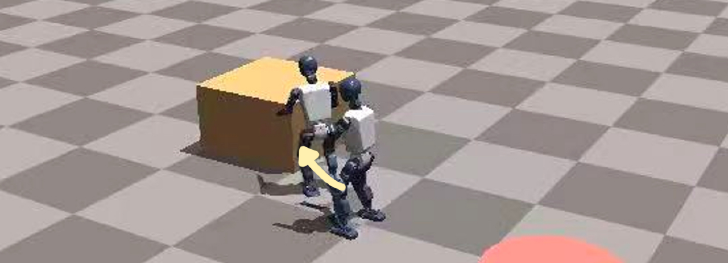}
        \caption{w/o Align Reward}
        \label{fig:wo_align}
    \end{subfigure}
    \caption{\textbf{Object-Goal Alignment Reward Ablation.} The alignment reward enables the policy to reposition around the object before pushing it toward the goal (a). Without it, the robot tends to make premature contact and fails to push the object to the goal (b).}
    \label{fig:align_reward_ablation}
\end{figure*}

\subsection{Performance Across Initial Configurations and Masses}
\label{sec:performance_quantitative_simulation}

We conduct two sets of quantitative simulation experiments using three metrics: \textit{Success Rate (SR)}, \textit{Terminated Pre-Contact (Pre-C.)}, and \textit{Terminated Post-Contact (Post-C.)}. An episode is successful if the object is pushed within 0.3\,m of the goal and remains there for 2\,s. Pre-C. and Post-C. denote that the robot falls before and after object contact, respectively.

As shown in~\tabref{object_goal_configuration_exp}, we evaluate three goal-position categories based on the relative goal angle $\theta$: \textit{front} ($|\theta| \in [0^\circ, 60^\circ]$), \textit{lateral} ($|\theta| \in [60^\circ, 120^\circ]$), and \textit{rear} ($|\theta| \in [120^\circ, 180^\circ]$). Performance remains consistently high across all configurations, with success rates above 97\% for the teacher and 90\% for the student. Even in rear configurations, which require longer-horizon repositioning, performance degrades only marginally.

As shown in~\tabref{object_mass_configuration}, we evaluate four mass ranges: \textit{light} (1--3\,kg), \textit{medium} (3--5\,kg), \textit{heavy} (5--8\,kg), and \textit{extra-heavy} (8--12\,kg), with the last range outside the training distribution. The policy remains robust across all in-distribution masses and achieves a 69\% success rate in the extra-heavy setting.

Across these experiments, we observe three failure modes: failing to push the box to the target, falling before contact, and falling after contact. Compared with the teacher, the student exhibits more target-reaching failures and post-contact falls, highlighting the added challenge of replacing privileged object-state information with noisy onboard visual observations while maintaining closed-loop performance.

\begin{figure*}[t]
    \centering
    \begin{subfigure}[t]{0.48\linewidth}
        \centering
        \includegraphics[width=\linewidth]{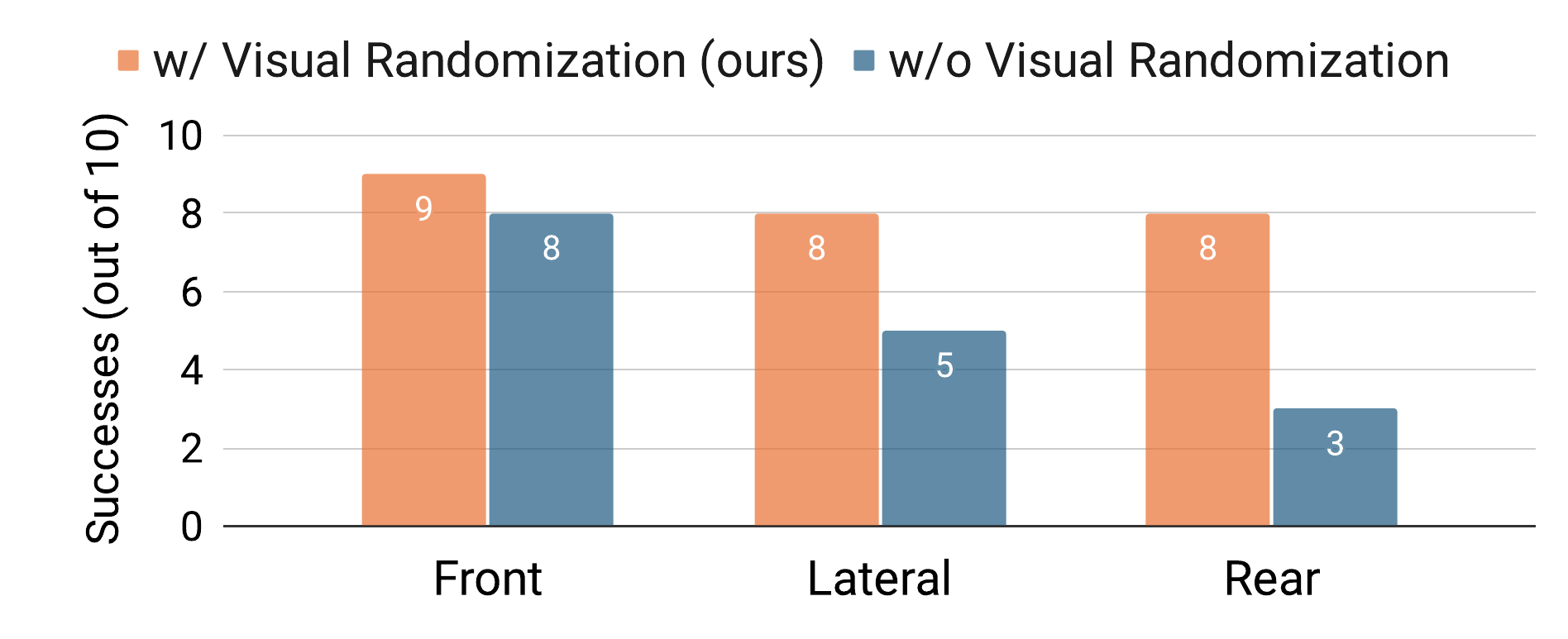}
        \caption{Different Goal Positions}
        \label{fig:diff_goal_pos}
    \end{subfigure}
    \begin{subfigure}[t]{0.48\linewidth}
        \centering
        \includegraphics[width=\linewidth]{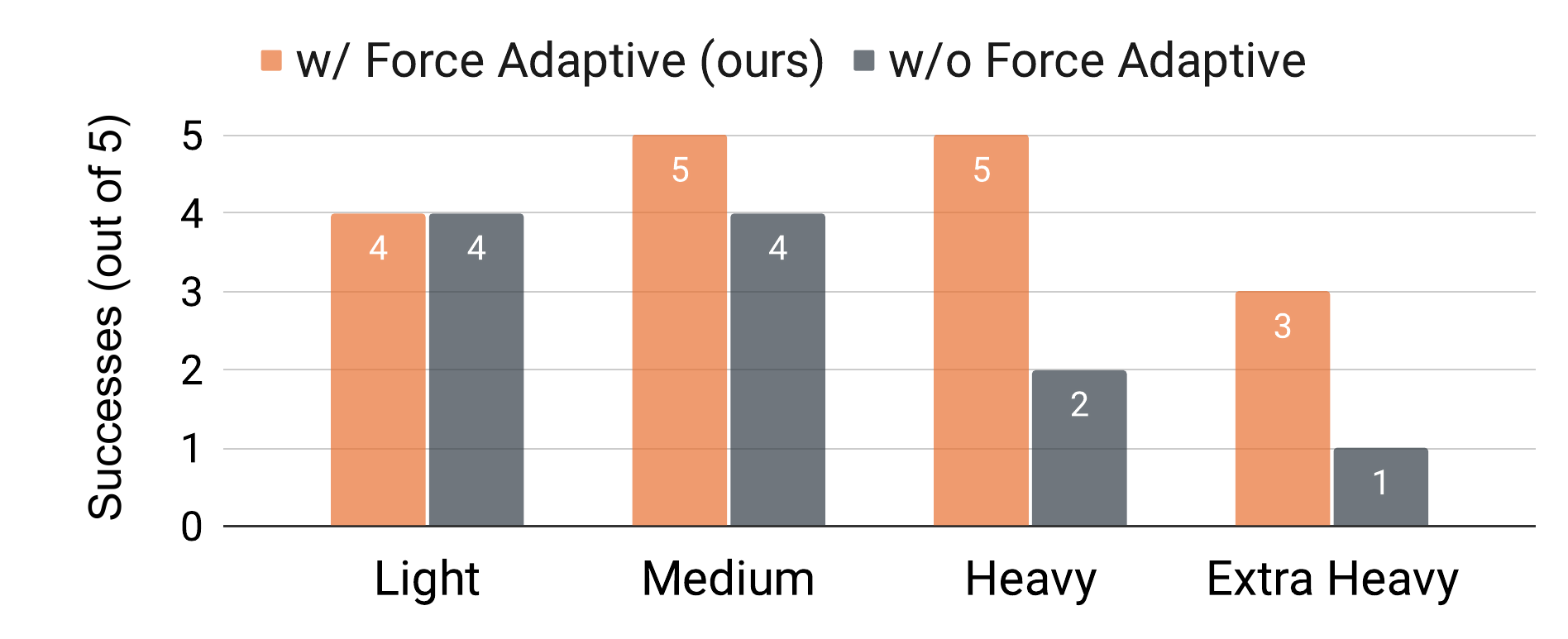}
        \caption{Different Object Masses}
        \label{fig:diff_mass}
    \end{subfigure}
    \caption{\textbf{Real World Deployment Results}. \policyname{} demonstrates strong sim-to-real transfer performance. Real-world experiments highlight the benefits of visual randomization across different goal positions (a) and the force-adaptive low-level controller across varying object masses (b).}
    \label{fig:real_world_result}
\end{figure*}

\begin{figure*}[t]
    \centering
    \begin{subfigure}{\linewidth}
        \centering
        \includegraphics[width=\linewidth]{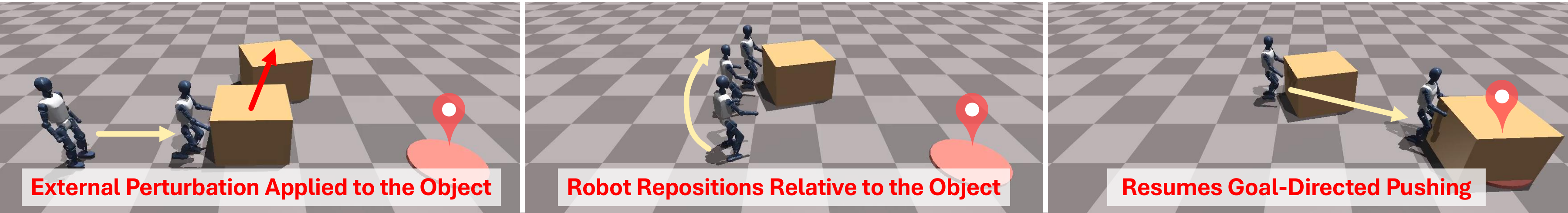}
        \caption{External perturbation recovery.}
        \label{fig:perturbation}
    \end{subfigure}
    
    \vspace{0.5em}
    
    \begin{subfigure}{\linewidth}
        \centering
        \includegraphics[width=\linewidth]{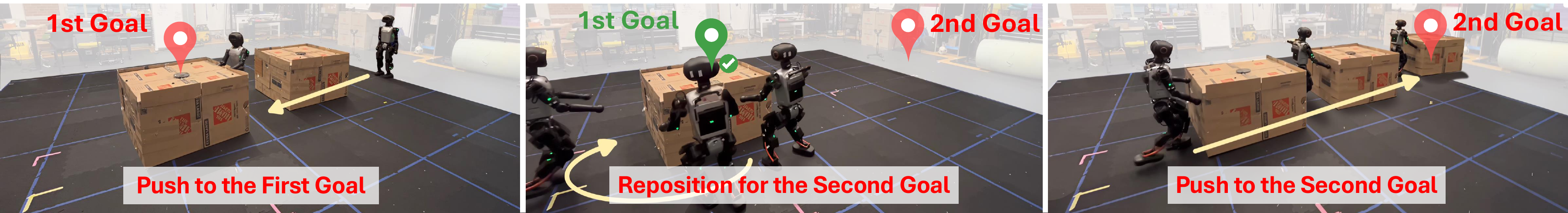}
        \caption{Sequential multi-goal pushing.}
        \label{fig:long_horizon}
    \end{subfigure}
    
    \caption{\textbf{Closed-Loop Demonstration.} The policy reacts to external disturbances (a) and supports sequential goal switching without reset (b), illustrating closed-loop control.}
    \label{fig:closed_loop}
\end{figure*}

\subsection{Effect of Force-Adaptive Whole-Body Control on \policyname{}}
\label{sec:effect_of_fa_control}

To evaluate the impact of force-adaptive whole-body control, we train ablated variants of \policyname{} by disabling random end-effector force perturbations during FALCON pretraining. We evaluate both teacher and student policies under varying initial goal configurations and object masses using the same setup as \secref{performance_quantitative_simulation} (\tabref{object_goal_configuration_exp} and~\tabref{object_mass_configuration}). Across all settings, force-adaptive control yields higher success rates and fewer terminations. Without force adaptation, performance degrades notably in lateral and rear configurations and for heavier objects, as the robot struggles to push the object to the target goal location. \figref{force_adaptive_ablation}~presents a qualitative comparison of the robot’s behavior with and without the force-adaptive controller. Without the force-adaptive controller, the robot struggles to apply consistent contact forces and often resorts to impulsive kicking motions, leading to unstable behavior and an increased risk of falling.
Overall, these results show that force-adaptive whole-body control is beneficial for robust loco-manipulation in \policyname{}.

\subsection{Impact of Object–Goal Alignment Reward}
\label{sec:align_reward}

Without the object–goal alignment reward, we observe a consistent failure mode in which the robot prematurely approaches the object when the object lies on the opposite side of the robot from the goal as shown in \figref{align_reward_ablation}. In such cases, the policy is driven by the $R_\text{reach}$ reward to make early contact, which prevents effective pushing toward the goal. As a result, the policy struggles to explore behaviors that require first repositioning the robot to a more favorable location before pushing. In contrast, introducing the object–goal alignment reward encourages deliberate repositioning prior to contact, significantly improving task success in these challenging configurations by promoting long-horizon execution rather than myopic contact strategies.

\subsection{Sim-to-Real Transfer Performance}
\label{sec:sim2real_performance}

To evaluate sim-to-real transfer, we deploy the policy on the Booster T1 humanoid to push a 5kg cardboard box measuring approximately $1.0 \times 1.0 \times 0.7$m. During deployment, we use motion capture to provide the relative goal position. While this quantity could also be obtained from a separate localization system, using motion capture allows us to focus the evaluation on the proposed visual object-pushing system.
We first evaluate performance across three goal configurations: \textit{front} (goal at $0^\circ$), \textit{lateral} (goal at $\pm90^\circ$), and \textit{rear} (goal at $180^\circ$). For each configuration, we perform 10 real-world trials, with the object randomly placed within the robot’s initial field of view following the ranges used during training. As shown in~\figref{diff_goal_pos}, the policy achieves consistently strong real-world performance across all goal configurations. 
In addition, we evaluate robustness to object mass by attaching additional weights to the box. We perform 5 real-world trials for each of the four mass conditions: \textit{light} (no additional weight), \textit{medium} (2.5kg), \textit{heavy} (4.5kg), and \textit{extra-heavy} (12kg). To isolate the effect of object mass, we fix the goal location 4m directly in front of the robot. 
As shown in~\figref{diff_mass}, the policy equipped with force-adaptive whole-body control maintains stable performance across all tested mass conditions, whereas the non–force-adaptive baseline degrades substantially as object mass increases. Notably, \policyname{} successfully pushes a 17kg object (12kg payload + 5kg box), exceeding half of the Booster T1 robot’s body weight, despite not being trained on such heavy objects in simulation. This real-world trend mirrors the simulation results, further validating the role of force-adaptive control in enabling robust sim-to-real transfer.

\subsection{Effectiveness of Visual Randomization}
\label{sec:effectiveness_visual_randomization}
To assess the effectiveness of the proposed visual randomization strategy, we conduct an ablation study by training a policy without visual randomization and evaluating it under the same real-world conditions described in the previous subsection (\figref{diff_goal_pos}). Without visual randomization, policy performance degrades substantially for tasks that require longer-horizon execution, particularly in lateral and rear goal configurations. As the robot moves through the environment, it encounters a wider range of background structures and viewpoints, increasing sensitivity to visual noise and domain mismatch. In the absence of visual randomization during training, these effects lead to unstable behaviors, including confusion between background objects and the target object or failure to correctly localize the goal. In contrast, visual randomization improves robustness to such visual variations, enabling more reliable sim-to-real transfer.

\subsection{Closed-Loop Object Pushing}
\label{sec:closed_loop_behavior}
To evaluate whether the policy performs closed-loop control during object pushing, we conduct three experiments in both simulation and the real world.

\paragraph{External Perturbation.}
As shown in~\figref{perturbation}, we apply a lateral impulse perpendicular to the pushing direction, displacing the object from its nominal path. The robot reacts by halting forward motion, reorienting to locate the displaced object, navigating around it to re-establish object–goal alignment, and subsequently resuming pushing toward the target. Successful task completion under this disturbance demonstrates online feedback correction rather than open-loop execution.

\paragraph{Off-Centered Mass Distribution.} 
As shown in~\figref{first_figure}d, we attach additional weights to one side of the object to create an off-centered center of mass (COM). This causes the box to rotate and deviate from the desired trajectory when being pushed. Similar to the external-perturbation example in~\figref{perturbation}, the robot detects this deviation and repositions itself to push the box to the goal.

\paragraph{Sequential Goal Switching.} 
As shown in~\figref{long_horizon}, we evaluate reactive re-planning by assigning multiple goals sequentially. The robot first pushes the object to a front goal location. Upon reaching this target, a new goal is issued behind the robot. The robot repositions itself by navigating around the object and then pushes it toward the updated goal position without reset.

Collectively, these experiments qualitatively demonstrate that the policy operates in a closed-loop manner: actions are continuously conditioned on updated observations, enabling disturbance rejection, dynamic re-alignment, and goal adaptation during object pushing.

\vspace{-0.5em}
\section{Conclusion, Limitations, and Future Work}

In this work, we present \policyname{}, a visual goal-conditioned
humanoid object pushing system. By integrating a force-adaptive whole-body controller with a depth-based visuomotor policy trained via teacher–student distillation and extensive domain randomization, \policyname{} achieves robust closed-loop performance under diverse goal configurations and object physical properties. Extensive simulation and real-world experiments demonstrate strong generalization across goal positions, robustness to object masses, and effective sim-to-real transfer.
Despite these results, our approach has several limitations. First, we focus on single-object scenarios and do not explicitly consider environments with multiple objects or highly diverse background variations, which may introduce additional perceptual ambiguity. Second, the system assumes access to the relative goal position during deployment. Although this information could be supplied by a localization system, integrating such a system would introduce additional complexity. In future work, we aim to extend \policyname{} to more cluttered multi-object environments and develop an end-to-end framework that jointly reasons about object and goal localization from onboard perception.

\bibliographystyle{IEEEtran}
\bibliography{references}

\clearpage

\appendix

\begin{table}[H]
\centering
\footnotesize
\begin{tabular}{l r}
\toprule
\textbf{Actor observations} & \textbf{Dimensions} \\
\midrule
EE-object relative position & 9 \\
Object rotation matrix & 3 \\
Arm dof position & 10 \\
Arm dof velocity & 10 \\
Base linear velocity & 3 \\
Base angular velocity & 3 \\
Projected gravity & 3 \\
Relative goal position & 2 \\
Last action & 13 \\
\midrule
Total dim & 56\\
\midrule
\textbf{Priv. critic observations} & \textbf{Dimensions} \\
\midrule
Object contact & 1 \\
Object CoM & 3 \\
Object mass & 9 \\
Object dimension & 9 \\
Object linear velocity & 3 \\
Object angular velocity & 3 \\
\midrule
Total dim & 84 \\
\bottomrule
\end{tabular}
\caption{Observation dimensions for an asymmetric actor-critic setup used to train the high-level teacher policy.}
\label{tab:obs_dim}
\end{table}

\begin{table}[h]
\centering
\footnotesize
\begin{tabular}{l r}
\toprule
\textbf{Student proprioception observations} & \textbf{Dimensions} \\
\midrule
Projected gravity & 3 \\
Base angular velocity & 3 \\
Dof position & 23 \\
Dof velocity & 23 \\
Last action & 13 \\
Relative goal position & 2 \\
\midrule
Total dim & 67 \\
\midrule
\textbf{Student image observations} & \textbf{Dimensions} \\
\midrule
Depth image $(N_{\text{img}} \times H \times W)$ & $3 \times 32 \times 32$ \\
\bottomrule
\end{tabular}
\caption{Observation dimensions for the high-level student policy.}
\label{tab:obs_dim}
\end{table}

\begin{table}[h]
\centering
\footnotesize
\begin{tabular}{l l r}
\toprule
\textbf{Main Reward} & \textbf{Param. \textbf{$\sigma$}} & \textbf{Weight} \\
\midrule

Reach object & $\sigma_\text{reach} = 2.0$ & $1.25$ \\

Push object to goal
& $\sigma_\text{push} =10.0$ 
& $2.5$ 
 \\

Head track object
& $\sigma_\text{track} =0.5$ 
& $1.0$ 
 \\

Object-goal alignment
& $\sigma_\text{align} =0.25$
& $2.0$ 
 \\
\midrule
\textbf{Auxilary Penalty} & \textbf{Expression} & \textbf{Weight} \\

Command smoothness 
& $\|\mathbf{u}_t - \mathbf{u}_{t-1}\|_2^2$ 
& $-1.0$ 
\\

Object balance
& $
\|\theta_{\text{object tilt}}\|_2^2$ 
& $-0.5$ 
 \\

Upper joint action 
& $\|\mathbf{q}^\text{upper}_t\|_2^2$ 
& $-0.5$ 
 \\

\bottomrule
\end{tabular}
\caption{Reward terms and weights}
\label{tab:reward_terms}
\end{table}













\end{document}